\documentclass[11pt,a4paper]{article}

\usepackage[utf8]{inputenc}
\usepackage[T1]{fontenc}
\usepackage{lmodern}

\usepackage[margin=1in]{geometry}
\usepackage{setspace}
\usepackage{amsmath,amssymb,amsthm}

\usepackage{graphicx}
\graphicspath{{figures/}}
\usepackage{booktabs}
\usepackage{tabularx}
\usepackage{float}
\usepackage[section]{placeins}

\usepackage[numbers,square]{natbib}
\usepackage{hyperref}
\hypersetup{
  colorlinks=true,
  linkcolor=blue!60!black,
  citecolor=blue!60!black,
  urlcolor=blue!60!black,
  breaklinks=true
}
\usepackage{cleveref}

\usepackage{xcolor}
\usepackage{verbatim}


\begin{document}

\title{AIPsy-Affect: A Keyword-Free Clinical Stimulus Battery\\
for Mechanistic Interpretability of Emotion in Language Models}

\author{Michael Keeman\\
Keido Labs\\
\texttt{michael@keidolabs.com}}

\date{}

\maketitle


\begin{abstract}
Mechanistic interpretability research on emotion in large language models---linear probing, activation patching, sparse autoencoder (SAE) feature analysis, causal ablation, steering vector extraction---depends on stimuli that contain the words for the emotions they test.
When a probe fires on ``\emph{I am furious}'', it is unclear whether the model has detected anger or detected the word \emph{furious}.
The two readings have very different consequences for every downstream claim about emotion circuits, features, and interventions.
We release \textbf{AIPsy-Affect}, a 480-item clinical stimulus battery that removes the confound at the stimulus level: 192 keyword-free vignettes evoking each of Plutchik's eight primary emotions through narrative situation alone, 192 matched neutral controls that share characters, setting, length, and surface structure with the affect surgically removed, plus moderate-intensity and discriminant-validity splits.
The matched-pair structure supports linear probing, activation patching, SAE feature analysis, causal ablation, and steering vector extraction under a strong methodological guarantee: any internal representation that distinguishes a clinical item from its matched neutral cannot be doing so on the basis of emotion-keyword presence.
A three-method NLP defense battery---bag-of-words sentiment, an emotion-category lexicon, and a contextual transformer classifier---confirms the property: bag-of-words methods see only situational vocabulary, and a contextual classifier detects affect ($p < 10^{-15}$) but cannot identify the category (5.2\% top-1 vs.\ 82.5\% on a keyword-rich control).
AIPsy-Affect extends our earlier 96-item battery~\citep{keeman2026-whethernotwhich} by a factor of four and is released openly under MIT license.

\end{abstract}

\section{Introduction}
\label{sec:introduction}

A probe trained on residual stream activations distinguishes ``\emph{I am absolutely furious about this}'' from a neutral control.
What has it detected?

Two readings are compatible with the data.
The model has formed an internal representation of emotional meaning.
Or the model has detected the word \emph{furious}.
Both are consistent with the result.
They are not consistent with each other, and they have very different consequences for everything that comes next---emotion circuits, emotion features, emotion subspaces, interventions on emotion-mediated behaviour.

This is not a hypothetical concern.
Recent mechanistic interpretability work on emotion in language models has converged on a small set of evaluation corpora---crowd-enVENT~\citep{troiano2022-appraisal}, GoEmotions~\citep{demszky2020-goemotions}, synthetic emotion stories~\citep{sofroniew2026-emoconcepts}, emotion-elicitation prompts~\citep{wang2025-dollmfeel, chen2025-personavector}, and acted-speech benchmarks~\citep{zhao2026-audio}.
Every one of them marks the target emotion lexically in the stimulus.
The broader NLP emotion-classification landscape converges on the same assumption: a recent large-scale aggregation, SuperEmotion~\citep{deFortuny2025-superemotion}, explicitly motivates its choice of Shaver's six-emotion taxonomy on the grounds that it is ``\emph{lexically grounded---built from natural language emotion terms---which aligns well with the input modality of most NLP systems.}''
That alignment is precisely the assumption keyword-free stimuli are designed to test.

The literature is increasingly aware of this.
Tak et al.~\citep{tak2025-mechanistic} report that mid-layer attention units ``consistently attend to emotionally loaded tokens''---a description operationally indistinguishable from keyword routing.
Sofroniew et al.~\citep{sofroniew2026-emoconcepts} caution that probes built from synthetic emotion stories ``may be biased toward stereotypical or explicit expressions of emotion.''
Wang et al.~\citep{wang2025-dollmfeel} frame their own work as a deliberate move from emotion \emph{recognition} to emotion \emph{generation}, on the grounds that recognition probes a model's response to keyword-laden text rather than its internal affective computation.
None of these acknowledgments is a defect of the underlying papers.
They reflect the absence of a stimulus resource that would let the field test the keyword-independence assumption directly.

In our previous work~\citep{keeman2026-whethernotwhich}, we introduced Set~B: 96 third-person clinical vignettes designed to evoke specific emotions through situational and behavioural cues, with emotion vocabulary systematically removed, paired with 96 matched neutral controls.
The result was a clean dissociation.
Across six models, binary affect-detection probes reached AUROC 1.000 on the keyword-free items, saturating in early layers.
The model knew what it was looking at without any emotion vocabulary present.
The corpus was small enough to make the point but too small for the analyses the point invited.
With only four vignettes per emotion--domain cell, permutation tests of representational geometry reached 80\% power only at cosine gaps $\geq 0.009$---while observed gaps ran 0.001--0.005.
The methodology was sound; the corpus could not carry the program.

This paper releases \textbf{AIPsy-Affect}, a 480-item battery that scales the keyword-free design four-fold and adds three structural features beyond the original Set~B: an intensity gradient (peak vs.\ moderate) for dose--response analyses; a discriminant-validity split of vivid emotion-free narrative for separating affect detection from narrative-richness detection; and a fully matched-pair structure across all 192 clinical items.
The dataset is released openly under MIT license, with a canonical citation, on Hugging Face.

Our contribution is methodological, not empirical.
We do not claim a new finding about how transformers process emotion.
We provide the stimulus resource that lets such findings be tested under a regime where keyword-spotting cannot account for the result.
Where prior work establishes that emotion representations exist, AIPsy-Affect lets the field ask the next question: \emph{exist as what?}

\section{The Keyword Contamination Problem}
\label{sec:keyword-contamination}

The standard mechanistic interpretability protocol for studying emotion has three steps.
Select a corpus of emotion-labeled text.
Extract internal activations.
Show that some structure in those activations recovers the emotion labels---through linear probes~\citep{tigges2023sentiment, tak2025-mechanistic, reichman2025-emotionswhere}, activation patching~\citep{tak2025-mechanistic, keeman2026-whethernotwhich}, steering vectors~\citep{chen2025-personavector, sofroniew2026-emoconcepts, wang2025-dollmfeel}, sparse autoencoder features, or causal ablation.
The recovered structure is then interpreted as the model's internal representation of emotion.

That interpretation depends on a stimulus assumption that is rarely stated and almost never tested: that the recovered structure tracks the \emph{emotion} in the stimulus rather than the \emph{vocabulary used to label it}.
When the stimuli contain emotion keywords---as crowd-enVENT, GoEmotions, contrastive emotion prompts, and acted-speech transcripts all do---these two readings make identical predictions on the available data.
A feature that activates on ``\emph{she was devastated}'' cannot be distinguished from a feature that activates on ``\emph{devastated}''.
Not without a stimulus in which the emotion is present and the keyword is not.

Two consequences follow.

First, claims of the form ``\emph{the model represents emotion X at layer L}'' are, on keyword-rich stimuli, formally compatible with the much weaker claim ``\emph{the model maintains a layer-L representation of words that name emotion X}''.
The stronger claim may be true.
The weaker claim is sufficient to produce the result.
The data do not separate them.

Second, the ambiguity propagates.
Causal interventions, behavioural steering, and alignment-relevant manipulations built on these findings inherit it.
When a steering vector for \emph{desperate} increases reward hacking~\citep{sofroniew2026-emoconcepts}, we want to know whether the intervention is operating on a representation of desperation, on the lexical neighbourhood of \emph{desperate}, or on both.
The keyword-rich stimulus regime does not let us ask.

The methodological response is well-established outside mechanistic interpretability.
Cognitive neuropsychology routinely separates processing of a construct from processing of its label: lexical decision tasks, semantic Stroop paradigms, and clinical assessment instruments all activate the construct without naming it.
Stimulus designs of this kind have been used in human emotion research for decades, including in the affective neuroscience literature on amygdala function~\citep{phelps2005-amygdala, ledoux2000-emocircuits} and in appraisal-based emotion theory~\citep{scherer2009-architecture}.
The principle transfers directly.
To test whether a model's internal representation tracks the emotion or the word, present the emotion without the word.

In our prior work~\citep{keeman2026-whethernotwhich}, we performed the first such test on a transformer.
Across six models---Llama-3.2-1B, Llama-3-8B, Gemma-2-9B, base and instruct---binary affect detection on 96 keyword-free clinical vignettes saturated at AUROC 1.000 in the early-to-mid layers, while 8-class emotion categorisation dropped 1--7\% relative to keyword-rich stimuli, with the deficit shrinking with scale.
The dissociation was robust enough to motivate a program of follow-on work.
The corpus was not large enough to support all of it.

AIPsy-Affect is the corpus that follow-on work needs.

\section{Dataset Construction}
\label{sec:dataset}

\subsection{Design space and units}
\label{sec:design-space}

AIPsy-Affect targets the eight primary emotions of Plutchik's wheel~\citep{plutchik2001-natureemotions}---rage, terror, grief, loathing, amazement, admiration, ecstasy, vigilance---instantiated across a grid of psychological \emph{domains} and rendered as short third-person vignettes.
We retain the Plutchik taxonomy from our prior work~\citep{keeman2026-whethernotwhich} for continuity with prior MI emotion work.
Alternative carvings---Ekman's six, dimensional valence--arousal, appraisal-based---are not represented and would require separate stimulus families.

A \emph{domain} is a class of psychological situation that can plausibly elicit any of the eight emotions.
Financial betrayal, for instance, supports rage (theft by a trusted person), grief (loss of an inheritance), terror (threat to a child's future), loathing (moral disgust at the perpetrator), and four positive variants through corresponding outcome reversals.
Six domains are used: financial betrayal, professional violation, medical or institutional failure, caregiving responsibility, inheritance and intergenerational legacy, and public witness.

The 6-domain $\times$ 8-emotion factorial was a deliberate choice.
With three or fewer domains, an apparent emotion-specific representation cannot be cleanly distinguished from a domain-specific one---a feature that fires on ``financial betrayal'' regardless of emotion type would read as an AFFECT feature in a 3-domain design and as a DOMAIN feature in a 6-domain design.
The expansion is what makes the distinction recoverable.

Each emotion--domain cell contains four vignettes.
The fourfold replication supports within-cell variance estimation and permutation tests at the cell level---the $N$ that we identified in our prior work~\citep{keeman2026-whethernotwhich} as the binding constraint on representational-geometry analyses.

\subsection{Splits}
\label{sec:splits}

\begin{table}[ht]
  \centering
  \caption{The four splits of AIPsy-Affect.}
  \label{tab:splits}
  \begin{tabularx}{\textwidth}{lrX}
    \toprule
    Split & $n$ & Description \\
    \midrule
    \texttt{clinical}        & 192 & Keyword-free peak-intensity vignettes: 8 emotions $\times$ 6 domains $\times$ 4 vignettes \\
    \texttt{neutral}         & 192 & Matched neutral controls, paired 1:1 with \texttt{clinical} via \texttt{matched\_control\_id} \\
    \texttt{moderate}        &  48 & Moderate-intensity vignettes: 8 emotions $\times$ 3 domains (1--3) $\times$ 2 vignettes \\
    \texttt{complex\_neutral} &  48 & Vivid emotion-free third-person narrative for discriminant-validity testing \\
    \bottomrule
  \end{tabularx}
\end{table}

The \texttt{moderate} split intentionally covers domains 1--3 only.
We document the asymmetry rather than hide it.
The split exists to support intensity-gradient analyses by direct comparison against the peak \texttt{clinical} items in the same three domains.
A balanced 6-domain extension is planned as a separate release in the AIPsy dataset family rather than as a revision to this one.
Each item carries an \texttt{id}, target \texttt{emotion}, \texttt{intensity} (\texttt{peak}, \texttt{moderate}, or \texttt{none}), \texttt{domain} index, human-readable \texttt{domain\_label}, \texttt{matched\_control\_id} where applicable, \texttt{word\_count}, and \texttt{text}.

\subsection{Keyword-free construction}
\label{sec:keyword-free-construction}

The defining constraint on the \texttt{clinical} split is that no vignette contains a word that names its target emotion or any close synonym.
A vignette targeting rage may not contain \emph{furious}, \emph{outraged}, \emph{livid}, \emph{enraged}, \emph{seething}, \emph{incensed}, \emph{fuming}, or any affective adjective referring to anger; the same restriction applies to the other seven categories with their respective synonym sets.
The constraint extends to first-order morphological variants (\emph{fury}, \emph{furiously}) and to denotative phrases (\emph{lost his temper}).

Compliance was enforced through three layers of checking.
Each vignette was authored under the constraint by the designing clinical psychologist.
Each was then audited against an emotion-vocabulary lexicon assembled from the NRC Emotion Lexicon~\citep{mohammad2012-lexicon} and Plutchik-keyword lists.
Bag-of-words sentiment scoring (VADER,~\citealp{hutto2014-vader}) was used as a secondary check to surface high-valence affective adjectives the lexicon missed.
Flagged items were rewritten until the affect arose entirely from the situation described.

The validation results in~\cref{sec:validation} quantify the residual: zero emotion-label words appear in the keyword-free splits, and the high-valence words that \emph{do} appear are situational vocabulary---\emph{hand}, \emph{block}, \emph{clean}, \emph{approved}---rather than emotion vocabulary.

\subsection{Matched-pair construction}
\label{sec:matched-pair}

Each item in the \texttt{clinical} split has a counterpart in the \texttt{neutral} split, constructed by a deterministic transformation.
The transformation preserves the domain, the characters, the objects mentioned, the setting, the approximate word count, and the sentence structure of the source vignette.
It removes the elements that make the situation emotional---stakes, violation, threat, loss, moral transgression---and replaces them with their routine, expected, or mundane counterparts.

The neutral version is not a different scene.
It is the same scene with the affective valence stripped out.

This is the dataset's most important methodological asset.
Matched-pair contrast is the standard cognitive-neuropsychological response to surface-feature confounds: when the only systematic difference between two stimulus classes is the construct of interest, any internal representation that distinguishes them must, in principle, be tracking that construct rather than the surface.
A worked example follows.

\begin{quote}
\textbf{Clinical (\texttt{B-rage-d1-v1}).}\
\emph{The papers were spread across the floor where he'd thrown them. His daughter's college fund statement. Zero balance. His brother's signature on the withdrawal form---six separate transactions over nine months, each one just under the reporting threshold. The most recent was dated the same week his brother had sat at this table and told him he should put more money in for the girl's future. He picked up the phone. He put the phone down. He picked it up again.}

\medskip
\textbf{Matched neutral (\texttt{N-rage-d1-c1}).}\
\emph{The papers were spread across the table where he'd sorted them. His daughter's college fund statement. The balance had grown by four percent over the quarter. His brother had co-signed the original account---there were six deposits logged over nine months, each one automatic. The most recent was dated the same week they'd met for dinner. He picked up his phone to check the online portal, then set it aside and added the figures to the spreadsheet he'd been updating.}
\end{quote}

Same domain (financial, family).
Same characters (father, brother, daughter).
Same objects (papers, phone, fund statement).
Comparable length (82 vs.\ 81 words).
The systematic difference is purely situational: betrayal versus routine.

\subsection{Intensity gradient}
\label{sec:intensity-gradient}

The \texttt{moderate} split shares emotion--domain combinations with the \texttt{clinical} split for domains 1--3, but reduces the situational stakes.
A peak rage vignette presents a discovered betrayal with immediate, irreversible consequences.
The moderate counterpart presents the same kind of situation with ambiguity, temporal distance, or reduced severity---the kind of irritation that builds in a workday rather than the kind of fury that ends a relationship.

Calibration was done by the designing clinical psychologist.
No algorithmic intensity score was used; intensity in clinical work is a judgment, not a number.

The split supports dose--response analyses.
Under a detection-style account, internal representations of affect should scale with intensity.
Under a categorisation-style account, they should be approximately invariant within emotion category.
We do not test these predictions here.
We provide the stimulus support that would let them be tested.

\subsection{Discriminant-validity control (\texttt{complex\_neutral})}
\label{sec:complex-neutral}

The \texttt{complex\_neutral} split is a discriminant-validity control.
Users running binary affect probes on \texttt{clinical} versus \texttt{neutral} face a confound: the probe could be detecting \emph{narrative richness}---descriptive density, sensory detail, scene complexity---rather than affect.
The \texttt{complex\_neutral} split lets that confound be tested directly.

It contains 48 vivid third-person narratives matched to the \texttt{clinical} items on length, sentence complexity, and descriptive density, but with no human stakes, interpersonal dynamics, or moral dimensions: a lathe machining brass, a vessel navigating coastal waters, textile manufacturing processes.
A probe that fires on \texttt{clinical} items and on \texttt{complex\_neutral} items is detecting attentional or compositional richness.
A probe that fires on \texttt{clinical} but not \texttt{complex\_neutral} is detecting affect.

\subsection{Provenance and the AIPsy dataset family}
\label{sec:provenance}

AIPsy-Affect extends our Set~B~\citep{keeman2026-whethernotwhich} (8 emotions $\times$ 3 domains $\times$ 4 vignettes = 96 items, with 96 matched neutrals) by a factor of four on the keyword-free clinical split, adds the moderate intensity split, and extends the discriminant-validity neutral set from 24 to 48 items.
The original Set~B vignettes are retained as a subset of the released \texttt{clinical} and \texttt{neutral} splits, with identifiers preserved for cross-referenceability with prior work.

This release is intended as the first in a planned family of clinical-construct datasets---\emph{AIPsy}---built to support mechanistic interpretability research on psychological constructs beyond affect: emotion regulation, cognitive distortions, attachment, and other constructs from the clinical-psychology methodological tradition.
Future releases will follow the same design principles applied here.
Construct-driven stimulus design under explicit lexical constraints.
Matched-pair controls.
Intensity gradients where the construct admits them.
Discriminant-validity neutrals.

\section{Validation: A Three-Method NLP Defense Battery}
\label{sec:validation}

The keyword-free constraint is enforced by construction (\cref{sec:keyword-free-construction}).
But a constraint imposed at authoring time is only as good as its verification.
We validate the keyword-free property by asking, of every stimulus, what off-the-shelf natural language processing tools see in it.

The methods form a ladder of increasing sophistication: bag-of-words sentiment lexicon $\rightarrow$ bag-of-words emotion-category lexicon $\rightarrow$ contextual transformer classifier.
Each tests a complementary aspect of the keyword-free claim.
The convergent picture is the property the dataset is intended to provide.

A keyword-rich positive control is essential.
We constructed Set~A by sampling 90 keyword-rich vignettes from crowd-enVENT~\citep{troiano2022-appraisal} (10 per emotion across the eight Plutchik categories with a neutral filler).
Set~A is not part of the released AIPsy-Affect dataset.
It is curated from an existing public resource purely as a positive control for these validation analyses, and is described here for reproducibility.

\begin{figure}[!htbp]
  \centering
  \includegraphics[width=\textwidth]{fig_method_comparison}
  \caption{Three-method NLP defense battery applied to AIPsy-Affect.
  Left: VADER compound sentiment by split.
  Middle: NRC negative-emotion word fraction by split.
  Right: GoEmotions DistilRoBERTa $P(\text{emotional})$ by split.
  Set~A (keyword-rich, sourced from crowd-enVENT) is detected by all three methods; the keyword-free clinical split is detected as more emotional than its matched neutral controls only by the contextual classifier.}
  \label{fig:method-comparison}
\end{figure}

\subsection{Method 1 --- VADER (bag-of-words sentiment)}
\label{sec:vader}

VADER~\citep{hutto2014-vader} is a lexicon-and-rule-based sentiment analyser that scores text by combining valence values of recognised lexicon words.
We applied VADER to all 480 AIPsy-Affect items plus the 90-item Set~A control.

The positive control passes: Set~A is reliably more negative than the matched-neutral split (Mann--Whitney $U = 6{,}445$, $p = 5.80 \times 10^{-4}$).
On the critical contrast---\texttt{clinical} versus \texttt{neutral}---VADER also returns a significant difference ($U = 11{,}500.5$, $p = 1.76 \times 10^{-10}$, rank-biserial $r = 0.312$; matched pairs Wilcoxon $W = 2{,}602$, $p = 6.66 \times 10^{-15}$).
At face value, this would suggest keyword leakage.
Lexicon cross-referencing tells a different story.

Of the 441 high-valence VADER lexicon words ($|\text{valence}| \geq 1.5$) appearing in the keyword-free splits, \emph{zero} are emotion-label words.
The top-frequency contributors are situational vocabulary: \emph{hand} (40 occurrences), \emph{block} (23), \emph{clean} (16), \emph{like} (12), \emph{clear} (11), \emph{wrong} (10), \emph{death} (8), \emph{died} (8), \emph{approved} (8), \emph{ready} (7).
The signal VADER picks up is narrative content---financial-betrayal vignettes contain \emph{withdrawal} and \emph{violations}; medical-failure vignettes contain \emph{death} and \emph{died}---not emotion vocabulary.
The compositional asymmetry is visible in the negative-versus-positive decomposition: \texttt{clinical} \texttt{neg\_mean} (0.031) is roughly three times that of \texttt{neutral} (0.011), while \texttt{pos\_mean} is nearly identical (0.024 vs.\ 0.030).

Per-emotion VADER medians provide a sharper diagnostic.
If VADER were tracking the target emotion, positive emotions should score positive and negative emotions negative.
The actual pattern inverts in 6 of 8 categories: \emph{amazement} (positive) median $= -0.450$; \emph{admiration} (positive) $= -0.190$; \emph{vigilance} (neutral-positive) $= -0.226$; \emph{rage} (negative) $= +0.044$; \emph{grief} (negative) $= +0.058$; \emph{loathing} (negative) $= +0.070$.
VADER's signal is anti-correlated with the construct it would need to track for the significant difference to constitute keyword leakage.

The \texttt{moderate} split is not significantly different from \texttt{neutral} under VADER ($U = 3{,}847.5$, $p = 0.077$).
The \texttt{complex\_neutral} split is significantly different ($U = 2{,}972.5$, $p = 1.39 \times 10^{-4}$), for the same reason \texttt{clinical} is---vivid technical vocabulary triggers the lexicon (\emph{error}, \emph{block}) without referring to affect.

\subsection{Method 2 --- NRC Emotion Lexicon (bag-of-words, emotion-category-specific)}
\label{sec:nrc}

The NRC Emotion Lexicon~\citep{mohammad2012-lexicon} maps approximately 14{,}000 English words to Plutchik's eight emotion categories plus positive/negative valence.
Where VADER tests valence keyword leakage, NRC tests the more direct question: do emotion-\emph{category} words leak into the keyword-free splits?

The headline is that the keyword-free \texttt{clinical} split has a \emph{lower} NRC emotion-word hit rate (0.060) than its \texttt{neutral} counterpart (0.071), and a marginally higher negative-fraction (0.013 vs.\ 0.009; $U = 22{,}344.5$, $p = 2.62 \times 10^{-4}$) whose absolute magnitude makes the result substantively negligible---a difference of four words per thousand.
The keyword-free design constraint, applied with respect to a much narrower lexicon than VADER's, holds.

A within-\texttt{clinical} Kruskal--Wallis test asks the more demanding question: does NRC discriminate \emph{between} the eight target emotions in the keyword-free vignettes?
It does not.
Six of the eight NRC emotion categories (anticipation, disgust, fear, joy, sadness, surprise) show no significant differentiation across Plutchik categories.
The remaining two (anger, trust) reach only weak significance ($p = 0.025$ and $p = 0.042$, uncorrected).
NRC cannot tell the eight emotions apart from the keyword-free text alone.

\subsection{Method 3 --- GoEmotions DistilRoBERTa (contextual transformer classifier)}
\label{sec:goemotions}

The third tier of the ladder is a contextual classifier---\href{https://huggingface.co/j-hartmann/emotion-english-distilroberta-base}{Hartmann's GoEmotions DistilRoBERTa}, a transformer fine-tuned on the GoEmotions Reddit corpus~\citep{demszky2020-goemotions} for seven-class emotion classification.

GoEmotions reliably detects that \texttt{clinical} items are more emotional than their matched neutrals: $P(\text{emotional}) = 0.439$ versus $0.236$ ($U = 27{,}032$, $p = 2.63 \times 10^{-15}$, $r = 0.733$; matched pairs $W = 1{,}441$, $p = 3.43 \times 10^{-24}$, mean $\Delta = +0.20$).
The intensity gradient is also visible---\texttt{clinical} (0.439) $>$ \texttt{moderate} (0.287) $>$ \texttt{neutral} (0.236)---at a magnitude consistent with dose-dependent affect.
This is the desired result: the affect is genuinely present in the text, and a sufficiently capable model can recover it from situational semantics.

The classifier \emph{cannot}, however, recover \emph{which} emotion is present.
Top-1 emotion-classification accuracy on the keyword-free \texttt{clinical} split is 5.2\%---below the 14.3\% chance baseline for seven classes---versus 82.5\% on the keyword-rich Set~A control.
This reproduces the ``whether, not which'' pattern from our prior work~\citep{keeman2026-whethernotwhich} at the NLP-method level: a contextual classifier sees that something affective is present, but cannot identify the affective category from situation alone.

\begin{figure}[!htbp]
  \centering
  \includegraphics[width=\textwidth]{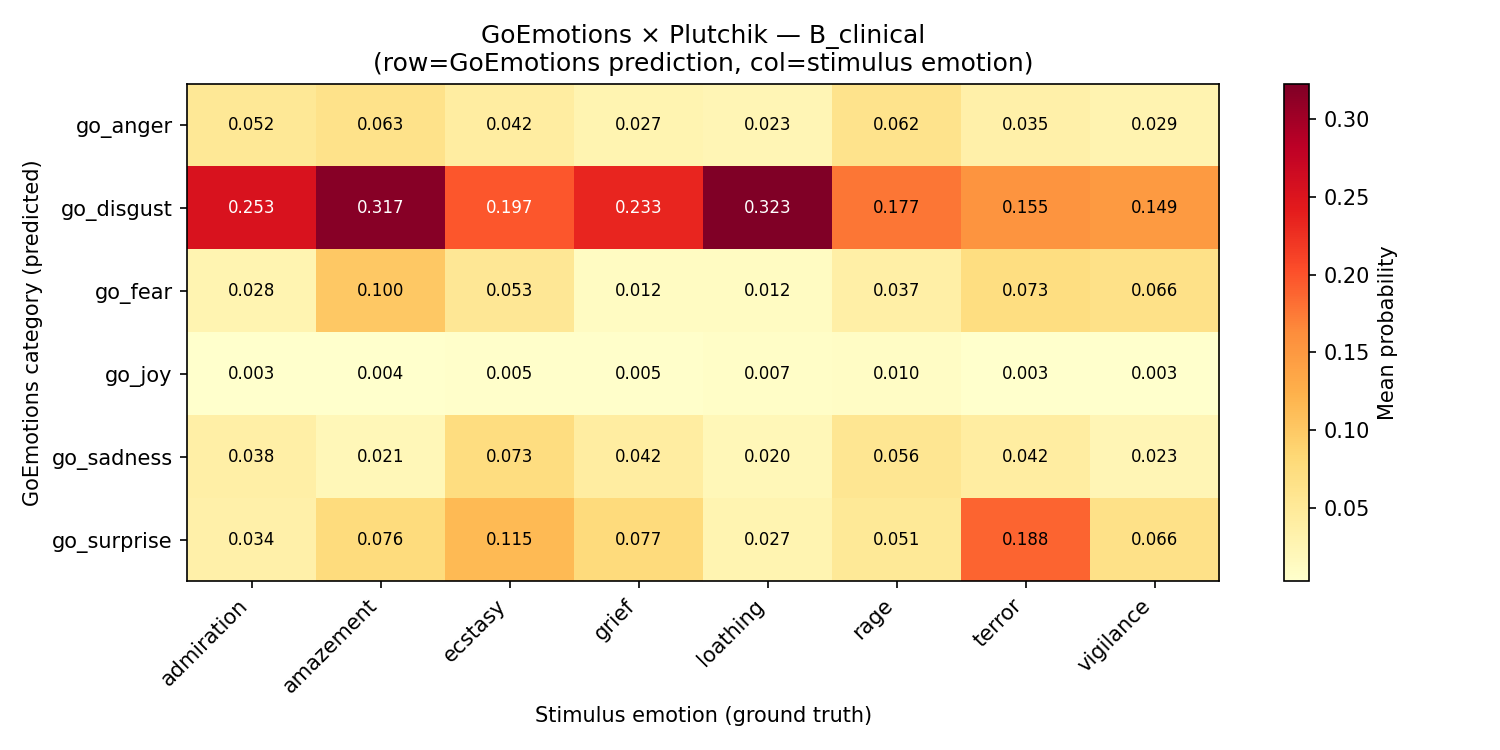}
  \caption{GoEmotions DistilRoBERTa class probabilities aggregated by target Plutchik emotion on the keyword-free \texttt{clinical} split.
  Probability mass collapses onto \texttt{go\_disgust} across most target emotions, indicating a generic ``something disturbing'' response rather than category-specific recognition.
  Top-1 accuracy is 5.2\% (chance $= 14.3\%$; 82.5\% on the keyword-rich Set~A control).}
  \label{fig:goemotions-heatmap}
\end{figure}

\subsection{Convergent interpretation}
\label{sec:convergent}

The three methods form a single coherent argument.
VADER, with its broad valence lexicon, detects a difference that is driven by situational vocabulary rather than emotion words; per-emotion medians are anti-correlated with target valence in 6 of 8 categories.
NRC, with its targeted emotion-category lexicon, returns no substantive difference at all---the keyword-free clinical vignettes contain \emph{fewer} emotion-category words than their neutral controls, and cannot be discriminated by emotion category.
GoEmotions, the contextual classifier, detects the affect ($p < 10^{-15}$) but fails to identify the category (5.2\% top-1 accuracy on keyword-free text vs.\ 82.5\% on keyword-rich text).

The affect in AIPsy-Affect is encoded in the situation, not in the vocabulary.
Bag-of-words methods that depend on emotion keywords cannot recover it.
A contextual model can detect \emph{that} it is present but not \emph{what} it is.
This is the property the dataset is built to provide: any internal representation that distinguishes a \texttt{clinical} item from its matched neutral is, by construction, not doing so on the basis of emotion-keyword presence.

\subsection{Known limitations of the validation}
\label{sec:limitations-validation}

Three caveats apply.

\emph{Loathing} carries the highest residual leakage risk in our authoring protocol.
Dehumanising language patterns---referring to a child as a ``scheduling problem''---may co-occur with disgust in transformer pretraining corpora through indirect lexical association, even when the explicit disgust vocabulary is absent.
The matched-pair design controls for this at analysis time, but downstream studies should flag loathing results for sensitivity analysis.

\emph{Positive-emotion neutrals} may differ from their clinical counterparts on dimensions other than affect.
Admiration and amazement scenarios are inherently engagement-heavy.
A probe firing on an admiration vignette and not on its matched neutral could in principle be detecting salience or interest rather than positive affect.
The \texttt{complex\_neutral} split is intended to support this kind of sensitivity analysis but does not by itself resolve it.

\emph{VADER's significant difference between \texttt{clinical} and \texttt{neutral}}---even though driven by situational rather than emotional vocabulary---means that compound sentiment scores are not matched across the two splits.
This does not affect the primary use case (mechanistic interpretability on internal representations), but researchers using VADER as a downstream behavioural metric should be aware of the asymmetry.

\section{Intended Use, Limitations, and a Worked Use Case}
\label{sec:use}

\subsection{Intended uses}
\label{sec:intended-uses}

AIPsy-Affect is designed for mechanistic interpretability research on emotion processing in language models.

The matched-pair structure of the \texttt{clinical} and \texttt{neutral} splits supports linear probing, activation patching, sparse autoencoder feature analysis, causal ablation, and steering vector extraction under a strong methodological constraint: any internal representation that distinguishes the two splits cannot be doing so on the basis of emotion-keyword presence.
The \texttt{moderate} split supports dose--response analyses against the peak \texttt{clinical} items in the same three domains.
The \texttt{complex\_neutral} split supports discriminant-validity tests separating affect detection from narrative-richness detection.
Behavioural evaluations---for instance, testing whether a model's response style varies with affective context in the absence of explicit emotional cues---are also supported by all four splits.

The dataset is \emph{not} designed for training emotion classifiers, for benchmarking general-purpose sentiment systems, or for clinical decision support.
It is a stimulus battery for hypothesis-driven research on internal model representations.

\subsection{A worked use case}
\label{sec:worked-use-case}

The proof-of-concept for this stimulus design appeared in our prior work~\citep{keeman2026-whethernotwhich}, which used the 96-item predecessor of the present \texttt{clinical} and \texttt{neutral} splits to test whether emotion processing in transformers is keyword-dependent.
Across six models---Llama-3.2-1B, Llama-3-8B, Gemma-2-9B; base and instruct---binary affect-detection probes trained on residual-stream activations achieved AUROC = 1.000 on the keyword-free items, saturating in early layers (9--38\% of normalised depth).
The same study reported that 8-class emotion categorisation dropped 1--7\% relative to keyword-rich stimuli, with the gap shrinking with scale---establishing affect reception as a keyword-independent computation distinct from emotion categorisation.

\begin{sloppypar}
The expanded dataset released here is intended to support the same class of analyses with the statistical power required for finer-grained tests: per-emotion feature specificity, intensity-dependent representational scaling, domain-versus-emotion variance decomposition, and discriminant-validity contrasts that the $n = 4$ per cell of the original could not resolve.
\end{sloppypar}

\subsection{Limitations}
\label{sec:limitations}

Several limitations are inherent to the dataset and should be carried forward into any work that uses it.

The dataset is English-only.
Cross-linguistic emotion processing research requires separate stimulus families.
The eight target emotions follow Plutchik's wheel of primary emotions~\citep{plutchik2001-natureemotions}; dimensional models (valence--arousal), Ekman's six, and appraisal-based taxonomies~\citep{scherer2009-architecture, troiano2022-appraisal} are not represented, and results obtained on AIPsy-Affect should not be extrapolated to taxonomies it does not test.

All vignettes were written by one clinical psychologist, the author.
This is mitigated by the multi-method NLP defense reported in~\cref{sec:validation} and by the open release of the corpus for independent inspection.
A multi-clinician replication remains a methodological priority and is planned as a parallel release in the AIPsy family.

The \texttt{moderate} split covers domains 1--3 only (48 items vs.\ 192 for the \texttt{clinical}/\texttt{neutral} splits).
Intensity-gradient analyses cannot be performed on domains 4--6 with the present release.
A balanced 6-domain extension is planned as a separate release.
There are no matched neutrals for the \texttt{moderate} split---binary affect-detection studies should use the peak \texttt{clinical}--\texttt{neutral} matched pairs, and peak-versus-moderate comparisons within emotion--domain cells provide a direct intensity contrast that does not require matched neutrals.

All vignettes are written in the third person.
First-person stimuli (``\emph{I found the papers on the floor\ldots}'') may activate different processing pathways and would require a separate stimulus family.

Finally, as noted in~\cref{sec:limitations-validation}, \emph{loathing} and the positive-emotion categories carry the highest residual confound risk and should be flagged for sensitivity analysis in any downstream study.

\section{Availability}
\label{sec:availability}

AIPsy-Affect is released under the MIT license at:

\begin{quote}
\url{https://huggingface.co/datasets/keidolabs/aipsy-affect}\\
DOI: \texttt{10.57967/hf/8215}
\end{quote}

The release includes the four splits described in~\cref{sec:splits}, full per-item metadata, and the dataset card with worked examples and design rationale.
The validation pipeline reported in~\cref{sec:validation} is reproducible from the released corpus with publicly available tools---VADER, the NRC Emotion Lexicon, and \href{https://huggingface.co/j-hartmann/emotion-english-distilroberta-base}{Hartmann's GoEmotions DistilRoBERTa classifier}.

To cite this work, please use this paper as a source:

\begin{verbatim}
@article{keeman2026-aipsyaffect,
  title         = {{AIPsy-Affect}: A Keyword-Free Clinical Stimulus
                   Battery for Mechanistic Interpretability of
                   Emotion in Language Models},
  author        = {Keeman, Michael},
  journal       = {arXiv preprint arXiv:2604.23719},
  year          = {2026},
  eprint        = {2604.23719},
  archivePrefix = {arXiv},
  primaryClass  = {cs.CL},
  doi           = {10.48550/arXiv.2604.23719},
  url           = {https://arxiv.org/abs/2604.23719}
}
\end{verbatim}


\FloatBarrier
\bigskip
\noindent\emph{Corresponding author: Michael Keeman (\texttt{michael@keidolabs.com}).}\\
\noindent\emph{Dataset available at \url{https://huggingface.co/datasets/keidolabs/aipsy-affect}.}

\clearpage
\bibliographystyle{unsrtnat}
\bibliography{bib}

\end{document}